\documentclass[10.5pt,twocolumn]{article}
\usepackage[left=2cm,right=2cm,top=2cm,bottom=2cm]{geometry}
\usepackage{authblk}
\usepackage{epsfig}
\usepackage{amssymb}
\usepackage{amsmath}
\usepackage{amsfonts}
\usepackage[font=bf]{caption}
\usepackage{subcaption}
\usepackage{multirow}
\usepackage{algorithm2e}
\usepackage{acronym}
\newacro{cnn}[CNN]{Convolutional Neural Network}
\newacro{dsp}[DSP]{Digital Signal Processing}
\newacro{hdl}[HDL]{Hardware Description Language}
\newacro{fpga}[FPGA]{Field-Programmable Gate Array}
\newacro{fifo}[FIFO]{First-In First-Out}
\newacro{gpu}[GPU]{Graphics Processing Unit}
\newacro{hadoc}[HADOC]{Hardware Automated Description Of CNNs}
\newacro{hls}[HLS]{High-Level Synthesis}
\newacro{moc}[MoC]{Model of Computation}
\newacroplural{moc}[MoC]{Models of Computation}
\newacro{ocr}[OCR]{Optical Character Recognition}
\newacro{qos}[QoS]{Quality of service}
\newacroplural{qos}[QoS]{Qualities of service}
\newacro{tpr}[TPR]{True Positive Rate}
\newacro{tdr}[TDR]{\ac{tpr} to \ac{dsp} Ratio}
\newacro{mnist}[MNIST]{Mixed National Institute of Standards and Technology database}
\newacro{usps}[USPS]{United States Postal Service}
\newacro{vip}[VIP]{Virtual Image Processor}
\newacro{simd}[SIMD]{Single Instruction on Multiple Data}
\newacro{vhdl}[VHDL]{VHSIC Hardware Description Language}

\usepackage[scaled]{helvet}

\usepackage[T1]{fontenc}

\begin{document}
\title{A Holistic Approach for Optimizing DSP Block Utilization of a CNN implementation on FPGA}
\author[1]{K.Abdelouahab}
\author[3]{C.Bourrasset} 
\author[1,2]{M.Pelcat}
\author[1]{F.Berry}
\author[4]{J.C.Quinton}
\author[1]{J.Serot}

\affil[1]{Institut Pascal,Clermont Ferrand, France}
\affil[2]{INSA, Rennes, France}
\affil[3]{CEPP Bull, Montpellier, France}
\affil[4]{Universit\'e Grenoble-Alpes,Grenoble, France}

\date{}
%\numberofauthors{6}
% \author{
% \alignauthor
% Kamel ABDELOUAHAB\\
%       \affaddr{Institut Pascal}\\
%       %%\affaddr{Universit\'e Blaise Pascal}\\
%       \affaddr{Clermont Ferrand, France}
% \alignauthor
% C{\'e}dric BOURRASSET\\
%       %%\affaddr{Institut Pascal}\\
%       \affaddr{CEPP Bull}\\
%       \affaddr{Montpellier, France}
% \alignauthor
% Maxime PELCAT\\
%       \affaddr{INSA}\\
%       %%\affaddr{Universit\'e Blaise Pascal}\\
%       \affaddr{Rennes, France}
% \and
% \alignauthor
% Fran\c cois BERRY\\
%       \affaddr{Institut Pascal}\\
%       %%\affaddr{Universit\'e Blaise Pascal}\\
%       \affaddr{Clermont Ferrand, France}       
% \alignauthor
% Jean-Charles QUINTON\\
%       %\affaddr{Institut Pascal}\\
%       \affaddr{Universit\'e Grenoble-Alpes, }\\
%       \affaddr{Grenoble, France}  
% \alignauthor
% Jocelyn SEROT\\
%       \affaddr{Institut Pascal}\\
%       %%\affaddr{Universit\'e Blaise Pascal}\\
%       \affaddr{Clermont Ferrand, France} 
% }

\maketitle

\begin{abstract}
Deep Neural Networks are becoming the de-facto standard models for image understanding, and more generally for computer vision tasks. As they involve highly parallelizable computations, \acp{cnn} are well suited to current fine grain programmable logic devices. Thus, multiple \ac{cnn} accelerators have been successfully implemented on FPGAs. Unfortunately, \ac{fpga} resources such as logic elements or \ac{dsp} units remain limited. This work presents a holistic method relying on approximate computing and design space exploration to optimize the \ac{dsp} block utilization of a \ac{cnn} implementation on \ac{fpga}. This method was tested when implementing a reconfigurable \ac{ocr} convolutional neural network on an Altera Stratix V device and varying both data representation and \ac{cnn} topology in order to find the best combination in terms of \ac{dsp} block utilization and classification accuracy. This exploration generated  dataflow architectures of 76 \ac{cnn} topologies with 5 different fixed point representation. Most efficient implementation performs 883 classifications/sec at 256 $\times$ 256 resolution using 8 \% of the available \ac{dsp} blocks.
\end{abstract}

%Deep Neural Networks are becoming the de-facto standard models for image understanding. As they involve highly parallelizable computations, CNN are well suited to current fine grain programmable logic devices. Thus, multiple CNN accelerators have been successfully implemented on FPGAs. Unfortunately, hardware resources such as logic elements or DSP units remain limited. This work presents an holistic method relying on approximate computing and design space exploration to find the best combination in terms of DSP utilization and classification accuracy. This method was tested when implementing a reconfigurable CNN on an Altera Stratix V device and varying both data representation and CNN topology.  Exploration generated  dataflow architectures of 76 CNN topologies with 5 different fixed point representation. Most efficient implementation performs 883 classifications/sec at 256 x256 resolution using 8 % of the available DSP blocks.

\section{Introduction}

\acf{cnn} techniques are taking part in an increasing number of computer vision applications. They have been successfully applied to image classification tasks \cite{hinton12,lecun98} and are wildly being used in image search engines or in data centers \cite{microsoft15}.
% A verifier otcharov, mais c'est du DL sur FPGA sur Datas center, je ne sais pas si c'est du image search par contre

\ac{cnn} algorithms involve hundreds of regular structures processing convolutions alongside non linear operations which allow \acp{cnn} to potentially benefit from a significant acceleration when running on fine grain parallel hardware.
This acceleration makes \acp{fpga} a well suited platform for \ac{cnn} implementation. In addition, \acp{fpga} provide a lower power consumption than most of the \acp{gpu} traditionally used to implement \acp{cnn}. It also offers a better hardware flexibility and a reasonable computation power, as recent \acp{fpga} embed numerous hard-wired \ac{dsp} units. This match motivated multiple state-of-the-art approaches \cite{neuflow}, as well as industry libraries \cite{alteraCNN15} focusing on \ac{cnn} implementation on \acp{fpga}.

Nevertheless, efficient \ac{cnn} implementations for \ac{fpga} are still difficult to obtain. \acp{cnn} remain computationally intensive while the resources of \acp{fpga} (logic elements and \ac{dsp} units) are limited, especially in low-end devices. In addition, \acp{cnn} have a large diversity of parameters to tweak which makes exploration and optimization a difficult task.

This work focuses on \ac{dsp} optimization and introduces a holistic design space exploration method that plays with both the \ac{cnn} topology and its fixed point arithmetic while respecting some \ac{qos} requirements. In summary, the contribution of this paper is threefold:
\begin{itemize}
	\item	A tool named Hardware Automated Description of \acp{cnn} (HADOC) is proposed. This utility generates rapidly dataflow hardware descriptions of trained networks.
    \item	An iterative method to explore the design space is detailed. An optimized hardware architecture can be deduced after monitoring performance indicators such classification accuracy, hardware resources used or detection frame-rate.
    \item	\acs{tpr}/\acs{dsp} metric is introduced. This ratio measures the quotient between classification performance of a \ac{cnn} (its \ac{tpr}) and the number of \acp{dsp} required for its \ac{fpga} implementation. This gives a quantitative measure of an implementation efficiency.

\end{itemize}

Thus, this paper is organized as follows: Section II summarises state-of-the art approaches for ConvNets implementations and optimization on \acp{fpga}. Section III provides \ac{cnn} background and links it to  dataflow \ac{moc}. Section VI introduces design space exploration for \acp{cnn} on \acp{fpga} and our method for holistic optimizing. Section V details exploration and implementation results. Finally, section VI concludes the papers.

%\newpage

\begin{figure*}[h]
	\centering
	\includegraphics[width=0.78\textwidth]{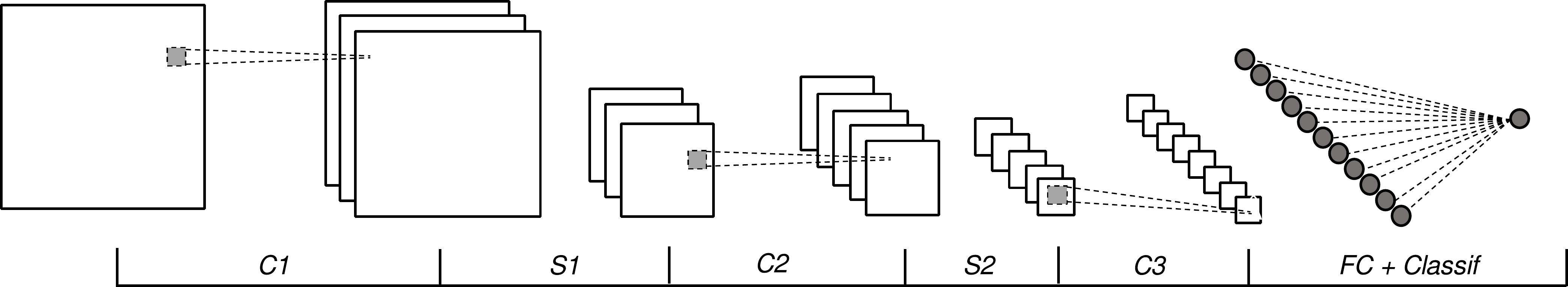}
	\caption{An example a \ac{cnn} topology,  3 convolutional layers interspersed with 2 sub-sampling layers and one fully connected stage}
    \label{dreamnet_img}
\end{figure*}

\section{Related work}
Neural networks are nowadays wildly used for computer vision tasks. Hence, multiple \ac{cnn} accelerators have been successfully implemented on \ac{fpga}. A non-exhaustive review of these can be found in \cite{lacey16}.

First attempt was in 1996 with \ac{vip} \cite{coutier96} : An \ac{fpga} based \ac{simd} processor for image processing and neural networks. However, since \acp{fpga} in that times were very constrained in terms of resources and logic elements, \ac{vip} performance was quite limited.

Nowadays, \acp{fpga} embed much more logic elements and hundreds of hardwired MAC operators (\ac{dsp} Blocks). State-of-the-art takes advantage of this improvement in order to implement an efficient feed-forward propagation of a \ac{cnn}. Based on \cite{lacey16}, and to our knowledge, best state-of-the-art performance for feed forward \ac{cnn} acceleration on an \ac{fpga} was achieved by Ovtcharov in \cite{microsoft15}, with a reported classification throughput of 134 images /second on ImageNet 1K \cite{hinton12}. Such a system was implemented on an a Stratix V D5 device and  outperformed most of state-of-the-art implementations such \cite{chakra10,peeman13,cnp10}.  Most of theses designs are \ac{fpga} based accelerators with a relatively similar architecture of parallel processing elements associated with soft-cores or embedded hardware processors running a software layer.

Regarding dataflow approaches for \ac{cnn} implementations, the most notable contribution was neuFlow \cite{neuflow}: A runtime reconfigurable processor for real-time image classification. In this work, Farabet and al. introduced a grid of processing tiles that were configured on runtime to build a dataflow graph for \ac{cnn} applications. It was associated to "luaFlow": a dataflow compiler that transforms a high-level flow-graph representation of an algorithm (in a Torch environment\cite{torch08})
into machine code for neuFlow. Such architecture was implemented on a Virtex 6 VLX240T and provided a 12 fps categorization for 512x375 images.

In \cite{zhang15}, an analytical design scheme using the roofline model and loop tiling is used to propose an implementation where the attainable computation roof of the FPGA is reached. This loop tilling optimization is performed on a C code then implemented in floating point on a Virtex 7 485T using Vivaldo HLS Tool. Our approach is different as it generates a purely dataflow architecture where topologies and fixed-point representations are explored.

\section{CNNs: background and implementation}
\subsection{Convolutional networks topology}
Convolutional Neural Networks, introduced in \cite{lecun98},  have a feed-forward hierarchical structure consisting of a succession of convolution layers interspersed with sub-sampling operators. Each convolution layer includes a large number of neurons, each performing a sum of elementary image convolutions followed by a squashing non-linear function (Figure~\ref{dreamnet_img}). A network topology can be described by its depth, the number of its neurons and their arrangement into layers. State-of-the-art \acp{cnn} for computer vision, such as \cite{hinton12}, are usually deep networks with more than 5 hidden layers and with thousands of neurons.

Numerous machine learning libraries \cite{torch08,theano12,caffe14} can be used to design, train and test \acp{cnn}. Caffe \cite{caffe14} is a C++ package for deep learning developed by the Berkeley Vision and Learning Center (BVLC).  This framework is leveraged on in this work as it benefits from a large community, contains Python and Matlab bindings, an OpenCL support and a "Model zoo repository'', i.e. a set of popular pre-trained models to experiment on. Moreover, \ac{cnn} topologies can easily be explored using Caffe.

Convolutional Neural Networks and more generally image stream processing algorithms can usually be expressed as sequences in an oriented graph of transformations. The \ac{cnn} layout matches intuitively a dataflow model of computation.

\begin{figure}[!h]
	\centering
	\includegraphics[width=0.4\textwidth]{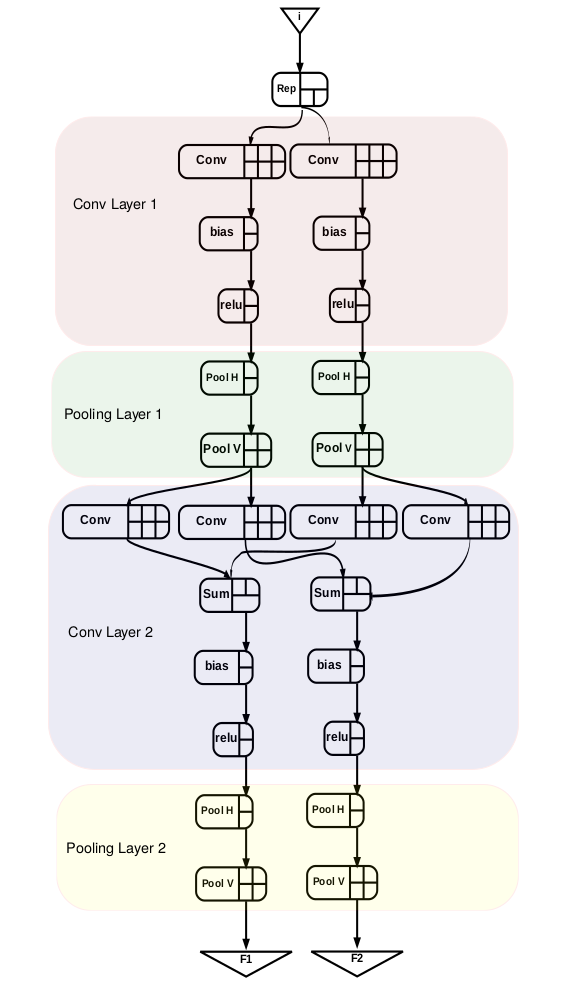}
	\caption{Dataflow graph of an elementary CNN with 4 layers of 2 neurons}
    \label{cnn_graph}
\end{figure}

\subsection{Dataflow \acp{moc} and \acp{cnn}}

The foundations of dataflow \acp{moc} were formalized by \cite{den74} in order to create an architecture where multiple fragments of instructions can process simultaneously a stream of data. Programs respecting dataflow semantics are described as a \textit{network} (graph) of fundamental processing units commonly called \textit{Actors} and communicating abstract data messages called \textit{tokens} on unidirectional \ac{fifo} channels. As each neuron applies convolutions with known kernels on streams of feature maps, a dataflow processing model can be appropriate for \acp{cnn} \cite{tensorflow15}. Thus, a high parallelism degree can be introduced at each layer on the network and the successive layers can be fully pipelined.
%\hl{cout }

As an example, Figure~\ref{cnn_graph} shows the dataflow graph of a simple feature extractor composed of two convolutional layers of two neurons each interspersed with pooling layers. Each actor in this graph performs elementary operations of a \ac{cnn} such as convolutions (\texttt{conv} actor), pooling (\texttt{poolV,poolH} actors), non linear activation (\texttt{ReLu} actor) and summation (\texttt{sum} actor). Such processing is done on a continuous stream \texttt{i} to extract two features \texttt{F1} and \texttt{F2} in the example.

% \subsection{Data representation and approximate computing}
% \newpage
\subsection{Implementation challenges on FPGAs}

The implementation of feed-forward propagation of a deep neural network in \acp{fpga} is constrained by the available computational resources.mendme Using floating-point to represent network parameters makes the convolution arithmetic very costly and requires many logic elements and \ac{dsp} blocks to be performed. Thus, \ac{cnn} implementations on \acp{fpga} usually build on fixed point representations of their data. Many studies have dealt with the aspects of deep learning with limited numerical precision \cite{hong2015,anwar15,suyog15}. The most common approach, such as in \cite{cnp10,gokhale14,neuflow}, consists of using a 16-bit fixed-point number representation which incurs little to no degradation in classification accuracy when used properly.

%In addition, floating point computations are commonly not supported in FPGA hardware \hlc{faux, c'est largement supporte mais c est juste que çà consomme beaucoup de ressources}. Thus, network parameters have to be represented in fixed point \hlc{have to - c'est un peu fort la quand meme}.
In this work, We explore different \ac{cnn} topologies and data representations. Theses parameters are adjusted to give the best trade-off between performance and resource consumption.

\section{Design space exploration}
\subsection{The \ac{tdr}}
Design Space exploration can be seen in our case as a method to deduce an efficient \ac{cnn} implementation that optimizes either classification accuracy (\ac{tpr}), hardware cost (with a focus in this paper on \ac{dsp} utilization), or a trade-off between these two elements. To measure this "trade-off", the \ac{tpr} to \ac{dsp} metric (\ac{tdr} in short) is introduced in equation \ref{tdr}. It computes the ratio between classification accuracy of the implementation (\ac{tpr}) and the number of instantiated \ac{dsp} blocks. \ac{tdr} can be seen as the amount of classification accuracy that a \ac{dsp} block contribute with. As an example, a \ac{tdr} of 0.4 \% means that every \ac{dsp} block brings 0.4 \% of classification accuracy. The higher the \ac{tdr} is, the more efficient the implementation will be.

\begin{equation}
    \label{tdr}
    \mbox {TDR(\%)} = \frac{\mbox {TPR(\%)} }{ \mbox{DSP}}
\end{equation}

This work aims to maximize the \ac{tdr} with a holistic approach that explores the \ac{cnn} topology and the data representation of the learned weights and biases. These two sets of parameters can be expressed as follows:\\
- The number of bits required to quantify network parameters (weights and biases)\\
- $N_1, N_2, ... , N_d$ : The number of neurons at each layer of the network.\\
% \begin{itemize}
%   \item $B$ : The number of bits required to quantify network parameters (weights and biases)
%   \item $N_1, N_2, ... , N_d$ : The number of neurons at each layer of the network.
% \end{itemize}
Other parameters, such as $d$ the depth of the network, or $k$ the size of the convolution kernels can also be adjusted, but this can only be done at a price of an increased exploration complexity. Thus, these parameters are left to future work. 

One way to explore the design space is by using an iterative method that generates the corresponding \ac{cnn} hardware architecture for each network topology and data representation size. The proposed design space exploration method is described with algorithm \ref{algo_explo}. The \ac{cnn} performances as well as the required hardware resources are monitored at each iteration. The implementation that maximizes the \ac{tdr} can be considered the one balancing the most application requirements and hardware constraints.

%In order to deduce the most efficient ConvNet architecture, design space should be explored. This exploration allows to pick the adequate number of bits required to quantify network parameters. In addition, this can determinate a suitable CNN topology to implement.

\begin{algorithm}
\caption{Design Space Exploration Procedure}
\label{algo_explo}
\ForAll{possible network topologies}
	{
	Train network\\
	\ForAll {possible fixed point representations}
    	{
        Generate hardware description\\
        Estimate classification accuracy \\
        Compute hardware utilization\\
        }
     }
\end{algorithm}

\subsection{The \acs{hadoc} tool}
The \ac{hadoc} utility is the tool proposed in this study for network exploration. Starting from a \ac{cnn} description designed and learned using Caffe, \ac{hadoc} generates the corresponding dataflow graph (Actors and \acp{fifo}) described as a Caph network. \ac{hadoc} also extracts the learned \ac{cnn} parameters from a caffe trained model (represented in a 32-bit floating point format) and quantizes the data into the desired fixed point-representation.

\begin{figure}[!h]
	\centering
	\includegraphics[width=0.31\textwidth]{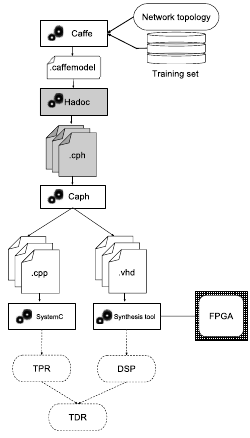}
	\caption{Conception flow of design space exploration  }
    \label{flow}
\end{figure}

Caph \cite{caph} is a dataflow language used here as an intermediate representation between the Caffe \ac{cnn} network and its hardware description in VHDL.  It is an image processing specific \ac{hls}  that interprets  a desired algorithm to create a digital hardware description that implements its behaviour. Compared to other \ac{hls} tools, Caph main feature is to generate a purely dataflow and platform independent architecture from a graph network. This architecture, described in \ac{vhdl} can be implemented using a synthesis tool. These tools indicates the used hardware resources (logic elements, memory blocks and hard-wired DSPs) of the \ac{fpga} target. Moreover, Caph provides a systemC back-end to perform a functional simulation of the designed hardware. This was used to estimate classification accuracy for each network topology and data representation scheme. Figure~\ref{flow} summarizes the conception flow and tools involved in this work. 

\section{Results}
This section describes an example of design space exploration method. \textit{Dreamnet}, a small convolutional neural network is explored, optimized and implemented on an Altera Stratix V 5SGSED8N3F45I4 device.
%his section introduces an example of design space exploration associated with %approximate computing in order to optimize CNN implementation on FPGAs. Therefore, %a basic convolutional neural network, \textit{Dreamnet}, was designed and %implemented on an Altera Stratix V 5SGSED8N3F45I4 device
%
%\hl{L'idee : En partant d'un reseau de profondeur 3 :  3 conv / 2 subsampling et 1 %FC:\\
%- Nombre optimal de neurons (best metric)\\
%- Dans quelle couche les mettre\\
%- Quelle precision }

\begin{figure}[h]
\centering
    \begin{minipage}[b]{0.5\textwidth}
    \centering
        \includegraphics[width=.68\linewidth]{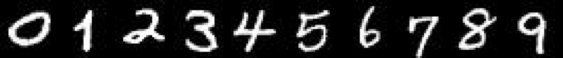}
        \subcaption{\ac{mnist}}
        \label{samples:mnist}       
    \end{minipage}

    \begin{minipage}[b]{0.5\textwidth}
    \centering
        \includegraphics[width=.68\linewidth]{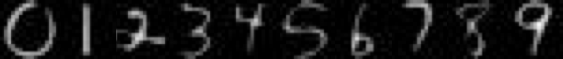}\\
        \subcaption{\ac{usps}}
        \label{samples:usps}
    \end{minipage}
    \caption{Differences between \ac{mnist} and \ac{usps} handwritten digits databases}
    \label{samples}
\end{figure}

\textit{Dreamnet} is a light \ac{cnn} designed for \ac{ocr} applications. It is inspired from the LeNet5 \cite{lecun98} as it includes 3 convolutional layers of 3x3 kernels interspersed with 2 sub-sampling stages that perform max-pooling operations. The depth of the network is the only constant parameter, the network topology (number of neurons per layer $N_{1}$ ,$N_{2}$, and $N_{3}$), and parameter representation $B$ being varied and explored. Table~\ref{dreamnet_tab} describes its topology. \textit{Dreamnet} is trained on 10000 images from the \ac{mnist} handwritten digit database of which a few samples are displayed in Figure~\ref{samples:mnist}. The classification accuracy of its implementation is then estimated on 1000 images of handwritten digits from the \ac{usps} database. The \ac{usps} database contains digits difficult to identify, as shown by samples in Figure~\ref{samples:usps}.

\begin{table}[!h]
\centering
    \begin{tabular}{|c|c|c|c|c|}
    \hline
    Layer    &  Size        & Operation             &   Kernel  \\ \hline
    C1       & $ N_1$       & Convolution           &   3x3     \\ \hline
    S1       & $ N_1$       & Max sub-sampling      &   2x2     \\ \hline
    C2       & $ N_2$       & Convolution           &   3x3     \\ \hline
    S2       & $ N_2$       & Max sub-sampling      &   2x2     \\ \hline
    C2       & $ N_3$       & Convolution           &   3x3     \\ \hline
    FC       &  10          & Inner product         &   -       \\ \hline
    Classif  &  10          & Softmax               &   -       \\ \hline
    \end{tabular}
    \caption{Dreamnet topology}
\label{dreamnet_tab}
\end{table}

In order to establish the optimal CNN topology, the space of possible configurations is explored. At each iteration, the tool chain of Figure~\ref{flow} is leveraged on. It consists of the caffe tool for specifying the network and learning parameters, the \ac{hadoc} tool to generate Caph code, the Caph compiler to generate VHDL, and finally the Altera Quartus II synthesizer to evaluate the required hardware resources.

For each topology, \textit{Dreamnet} is trained using caffe before generating the corresponding hardware. In this work, the choice is made to set,  in the most resource-hungry case, a limited number of 5 neurons for the first layer, 10 for the second layer, and 14 for the third layer. This topology is sufficient enough to offer reasonable classification accuracy for \ac{ocr} purpose on the \ac{mnist} database ({99.7} \% \ac{tpr} on the \ac{mnist} test-set). As a \ac{cnn} extracts features from an image hierarchically,  the number of neurons in a layer $N_i$ should be higher than the number of neurons in a previous layer $N_{i-1}$. Finally, a constant step $step = 2$ is chosen to iteratively increment the topology parameters. This step can be reduced to $1$ to have to have a more accurate optimization.
The other explored parameter is the data representation size $B$. On \textit{Dreamnet}, a data can have a maximum size of 7 bits. On one hand, this representation engenders a relatively low error rate compared to a floating-point reference. On the other hand, it prevents arithmetic over-flows to happen especially in the last stages of the network. In contrast, a minimum of 3 bits were used to represent the parameters which was the weakest precision usable to have acceptable classification rates. 

\begin{table}
\centering
    \begin{tabular}{|c|c|c|c|}
    \hline
    Parameter    & Description                      &  Min      &   Max         \\ \hline
    $ N_1$       & Number of neurons in C1          &  3        &   5           \\ \hline
    $ N_2$       & Number of neurons in C2          &  5        &   10          \\ \hline
    $ N_3$       & Number of neurons in C3          &  7        &   14          \\ \hline
    $ B $        & Representation size (in bits)    &  3        &   7           \\ \hline
    \end{tabular}
    \caption{Design space boundaries: Dynamics of the explored parameters}
\label{boundaries}
\end{table}

The design space boundaries being defined (summarized in table \ref{boundaries}), Algorithm~\ref{algo_explo} can be reformulated as Algorithm~\ref{detailed_explo}. These boundaries will lead to explore a total of 76 networks with 5 different data type representations (A total of 380 combinations). In order to estimate classification accuracy, SystemC processed the 10000 images of the test set at a rate of 66.6 classifications/second while the synthesis tool takes an average of 6 minutes to compute the number of required DSP blocks. Thus, an architecture is explored every 8.5 minutes with an Intel i7-4770 CPU.  
%Table \ref{sim_feat} gives average processing time for training, mapping and simulating the 380 architectures.

\begin{algorithm}[!h]
\caption{Design space exploration on \textit{Dreamnet}}
\label{detailed_explo}

\For{$  N_1 $ \textbf {in} $min(N_1)$ \textbf {to} $max(N_1)$ }
	{
	\For{$  N_2 $ \textbf {in} $  N_1 + step $ \textbf {to} $max(N_2)$}
    	{
        \For{$  N_3 $ \textbf {in} $  N_2 + step $ \textbf {to} $max(N_3)$}
    	    {
    	    \textbf{Caffe:}             Train network\\
    	    \textbf{for }{$B$ \textbf {in} $  min(B)  $ \textbf {to} $max(B)$}
    	    {\\
    	    \textbf{Hadoc + Caph :}     Generate hardware \\
    	    \textbf{SystemC:}           Simulate TPR\\
    	    \textbf{Quartus:}           Compute DSP utilization\\
    	    }
    	    }
        }
    }
\end{algorithm}

A few remarkable implementations are detailed in tab \ref{best}. The most efficient implementation considering \ac{tdr} is I1: it presents the best trade-off between hardware cost and classification accuracy. Table \ref{i1} gives post-fitting reports of I1. This architecture consumes low resources of the \ac{fpga}  as it uses 161 from the 1963 available \ac{dsp} blocks of the Stratix V device and 20 \% of the available logic elements while maintaining a 64.8\% classification rate on \ac{usps} at a rate of 57.93 MHz per pixel (which corresponds to 883 classifications per second with 256 $\times$ 256 image resolution). Therefore, I1 could be implemented on a low-end device with less logic resources and \ac{dsp} blocks. I2 is the implementation with the lowest number of neurons and data representation size, thus, this implementation has the weakest \ac{dsp} usage. Finally, we found that the configuration with the greatest classification accuracy on \ac{usps} is I3. This implementation is among the ones with the highest number of neurons and fixed-point representation, considering the design space boundaries established above.

%
%\newpage

\begin{table}[!h]
\centering
    \begin{tabular}{|c|c|c|c|c|c|c|c|c|}
        \cline{2-9}
        \multicolumn{1}{c|}{}        & \sc n1 & \sc n2 & \sc n3 & \sc B & \sc tpr \tiny{\ac{usps}} & \sc tpr \tiny{\ac{mnist}} & \sc dsp  & \sc tdr    \\ \hline
        \multicolumn{1}{|l|}{\sc i1}     & 4  & 6  & 8  & 5  & 64.8  \%         & 98.3 \%          & 161 & 0,40   \% \\ \hline
        \multicolumn{1}{|l|}{\sc i2}     & 3  & 5  & 7  & 3 & 48.7   \%         & 82.4 \%         & 140 & 0,34 \%   \\ \hline
        \multicolumn{1}{|l|}{\sc i3}     & 4  & 8  & 12  & 7  & 73.2  \%        & 99.7 \%          & 428 & 0.17 \%   \\ \hline
    \end{tabular}
\caption{Remarkable implementations}
\label{best}
\end{table}

\begin{table}[!h]
\centering
\begin{tabular}{|c|c|}

\hline
\sc Logic utilization \small{( in ALMs )}  &  53,779 / 262,400 ( 20 \% )   \\ \hline
\sc Total RAM Blocks   	           	&  109 / 2,567 ( 4 \% )  \\ \hline
\sc Total \ac{dsp} Blocks          &  161/1963 (8 \%) \\ \hline
\sc Frequency               		&  57.93 MHz\\ \hline
\sc Classification rate   \tiny{( at 256 $\times$ 256 )}  		&  883 frame/s   \\ \hline

\end{tabular}
\caption{{I\small{1}} implementation features on Stratix 5SGSED8N3F45I4 device}
\label{i1}
\end{table}

\subsection{Topology exploration}
\begin{figure}[!h]
	\centering
    \begin{minipage}[b]{0.5\textwidth}
    \centering
        \includegraphics[width=.72\linewidth]{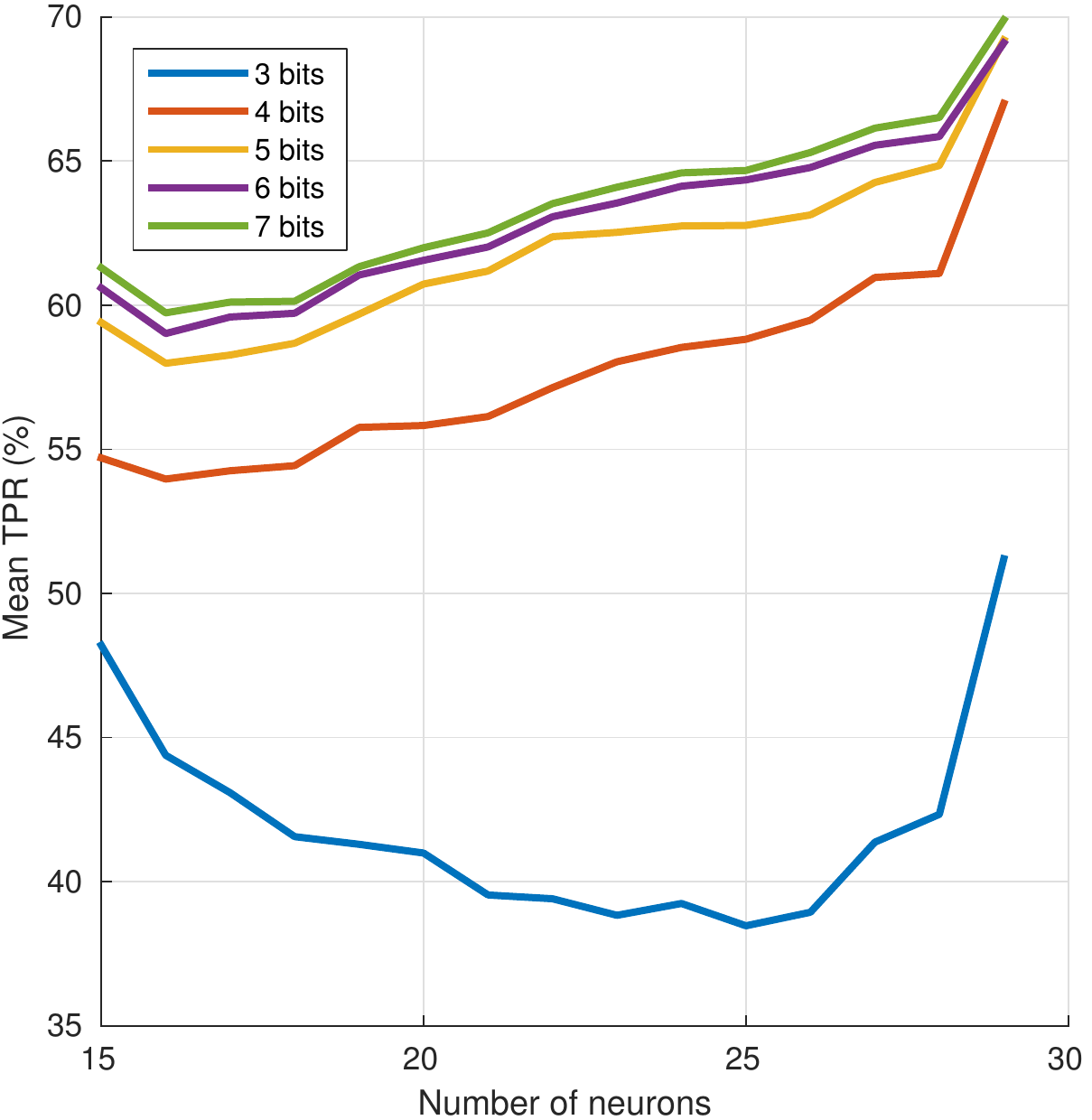}
        \subcaption{Classification accuracy}
        \label{tpr_ni}       
    \end{minipage}

    \begin{minipage}[b]{0.5\textwidth}
    \centering
        \includegraphics[width=.72\linewidth]{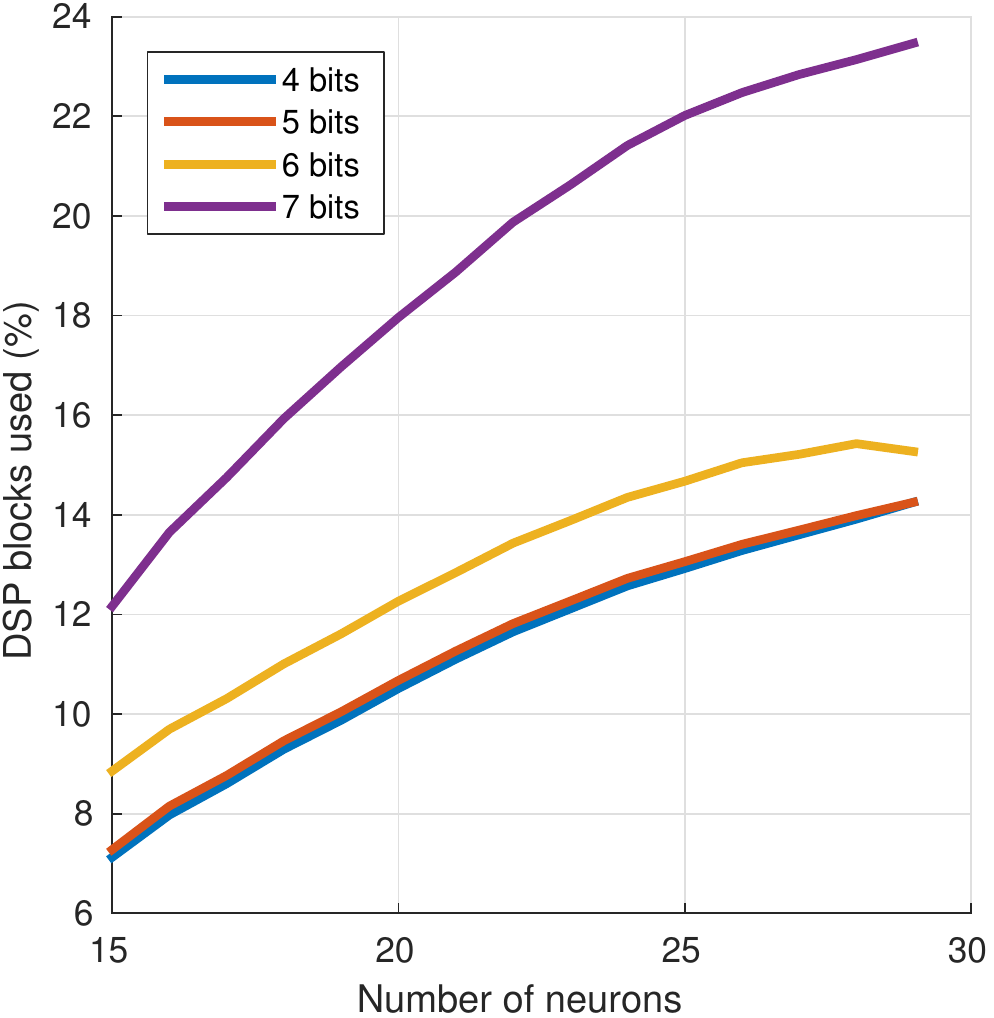}\\
        \subcaption{\ac{dsp} utilization}
        \label{dsp_ni}
    \end{minipage}
    
    \begin{minipage}[b]{0.5\textwidth}
    \centering
    	\includegraphics[width=0.76\textwidth]{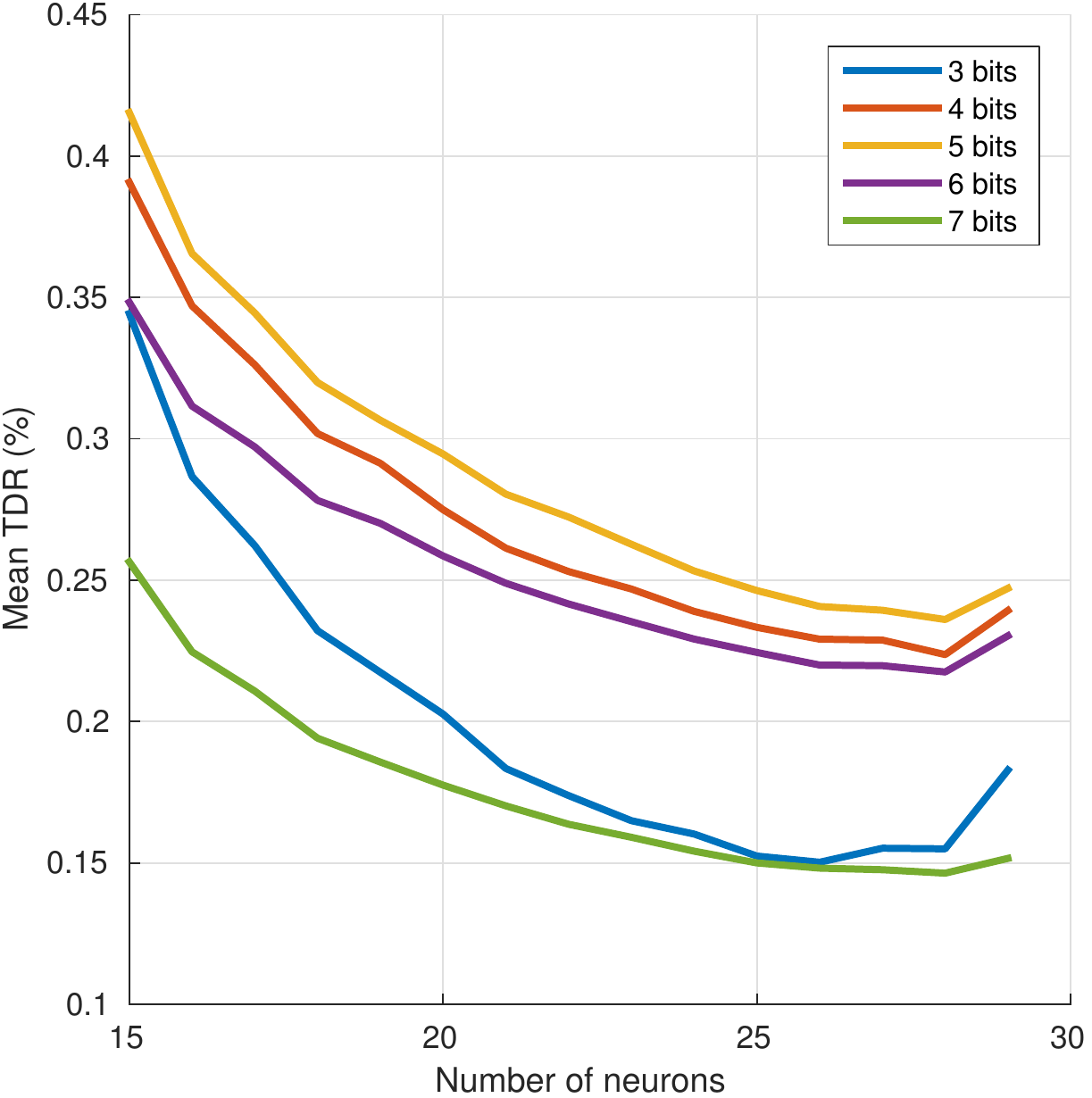}
    	\subcaption{Implementation efficiency }
		\label{tdr_ni}
    \end{minipage}
    
    \caption{Design space exploration for different \ac{cnn} topologies}
    \label{ni}
        
\end{figure}

When only network topology is explored (fixed-point representation maintained constant at 3,4,..,7 bits), we find that both of classification accuracy and \ac{dsp} utilization increase linearly with the number of neurons as shown in figures \ref{tpr_ni} and \ref{dsp_ni}.
This causes implementations efficiency to decrease as the number of neurons grows, as represented in figure \ref{tdr_ni}. The more "sized" a \ac{cnn} is, the less efficient its implementation will be.

\subsection{Data-representation exploration}
\begin{figure}
	\centering
    \begin{minipage}[b]{0.5\textwidth}
    \centering
        \includegraphics[width=.72\linewidth]{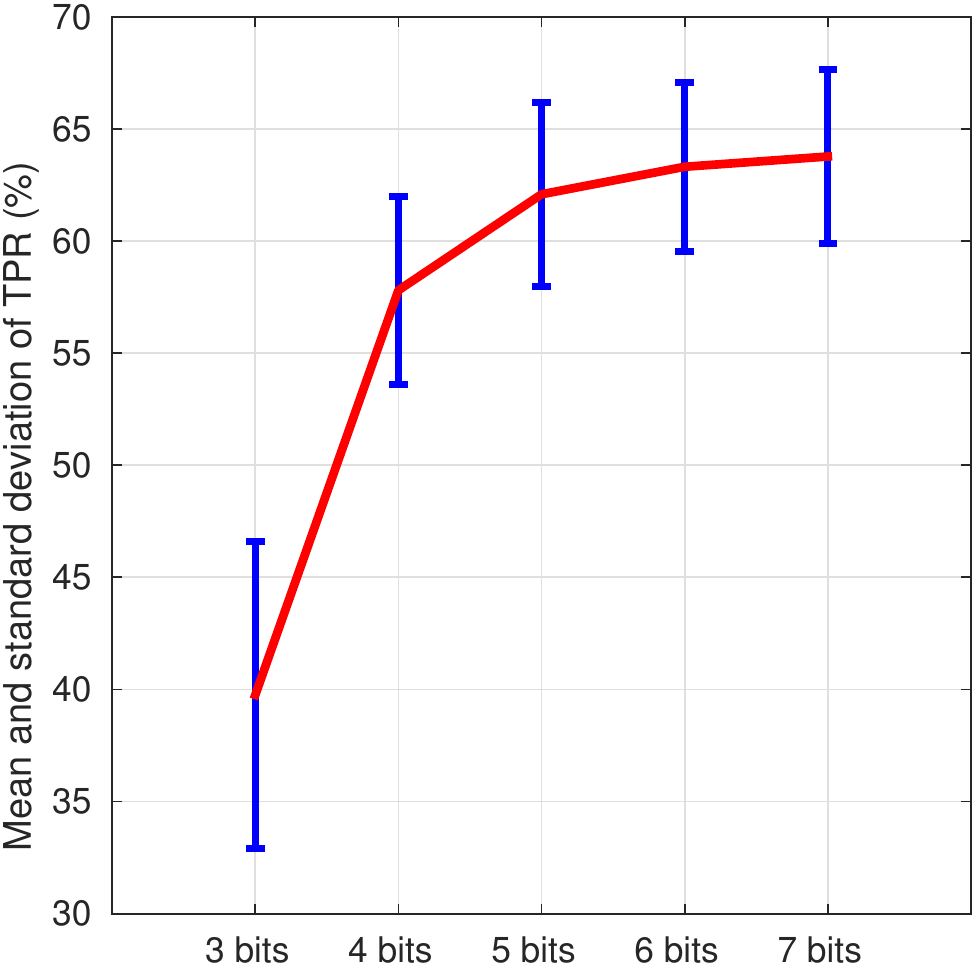}
        \subcaption{Classification accuracy}
        \label{tpr_nb}       
    \end{minipage}

    \begin{minipage}[b]{0.5\textwidth}
    \centering
        \includegraphics[width=.72\linewidth]{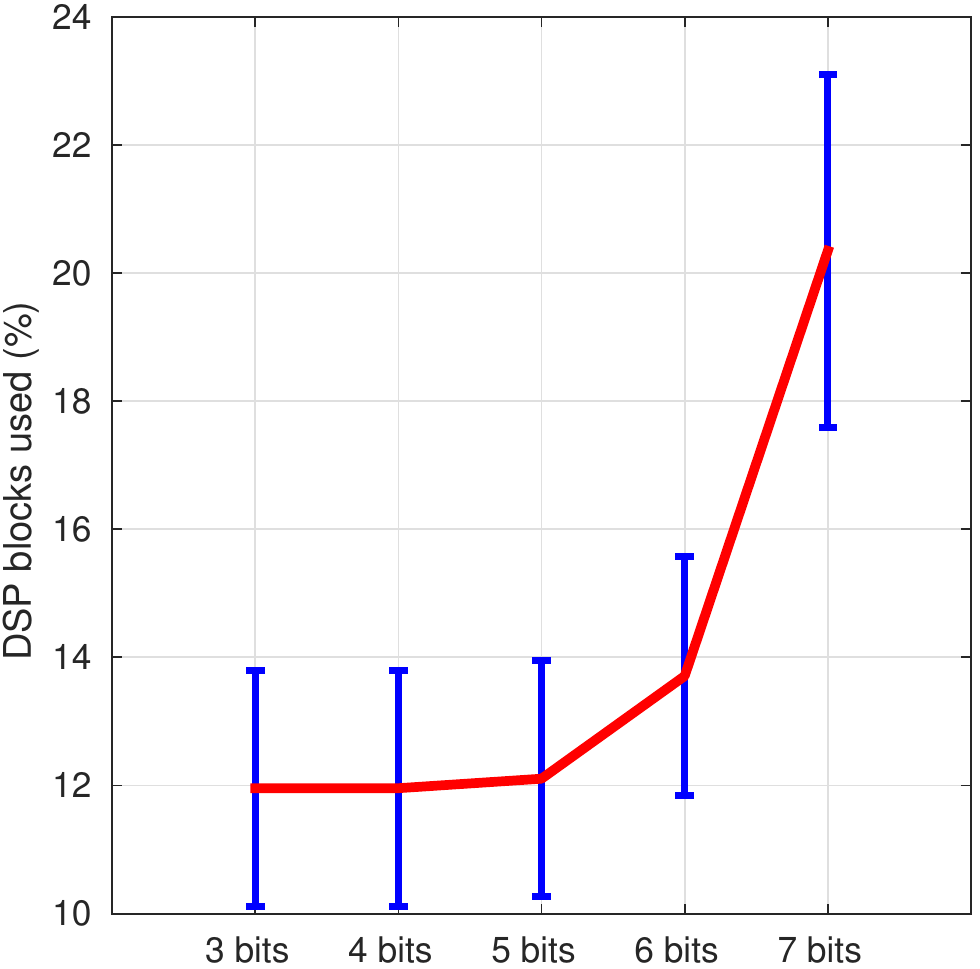}\\
        \subcaption{\ac{dsp} utilization}
        \label{dsp_nb}
    \end{minipage}

    \begin{minipage}[b]{0.5\textwidth}
    \centering
        \includegraphics[width=0.74\textwidth]{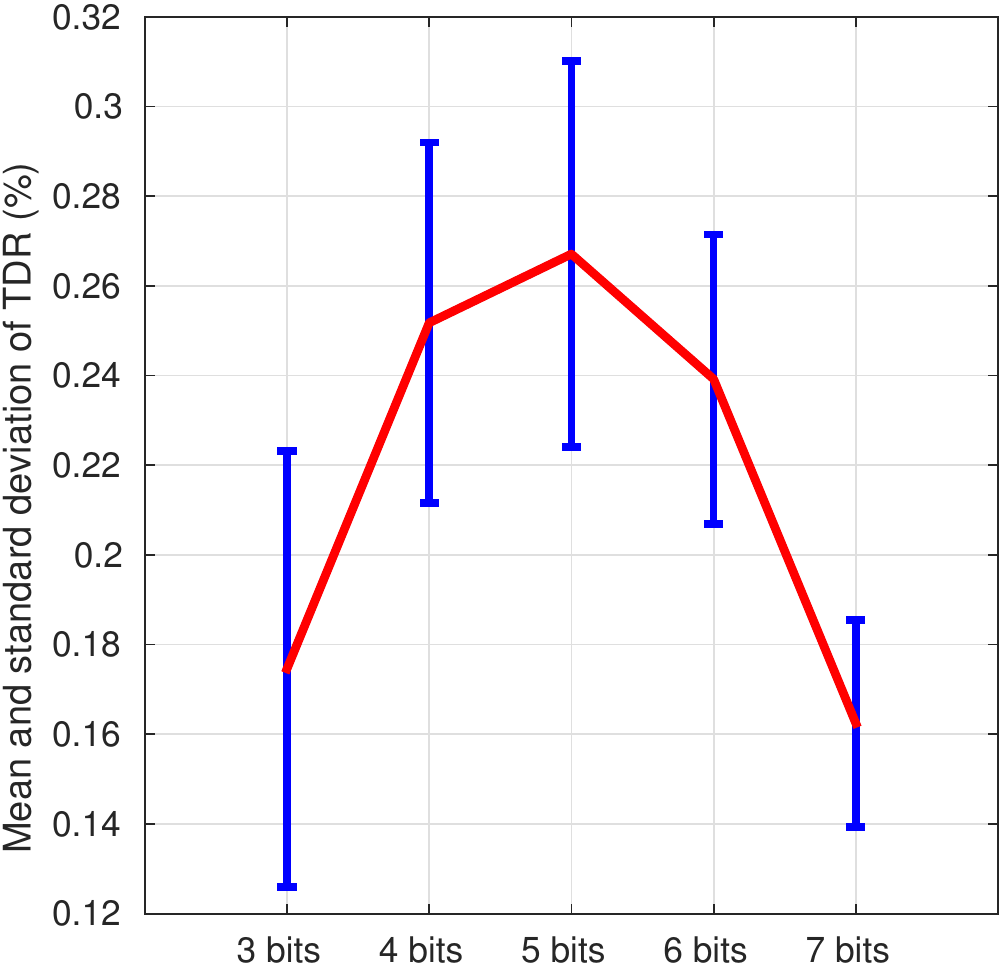}
    	\subcaption{Implementation efficiency}
		\label{tdr_nb}
    \end{minipage}
    
    \caption{Design space exploration for different representation size}
    \label{nbits}
\end{figure}

To see the approximate computing and numerical rounding effects on \acp{cnn} implementations, figures \ref{tpr_nb},\ref{dsp_nb} and \ref{tdr_nb} are plotted. They show the evolution of the average and standard deviation of network performances (in terms of \ac{tpr} and \ac{dsp} usage) for various data representations. 

It can be seen that the mean classification accuracy (for all explored topologies) grows with numerical precision. Moreover, figure \ref{tpr_nb} shows how a 5 bit representation can be sufficient enough to maintain tolerable classification accuracy. In addition, figure \ref{dsp_nb} shows that \ac{dsp} utilization grows quadratically when using different sizes of fixed point representations. Thus, the mean implementation efficiency (plotted in Figure \ref{tdr_nb}) have a maximum value (in our case 5 bits) that gives the best classification accuracy to \ac{dsp} utilization trade-off.

\subsection{Holistic Approach}
Table \ref{10best} presents the architectures with more than 70 \% classification accuracy on \ac{usps}. 5 of these implementations have a different network topology while 3 different data-representations are present. 

If only a topology exploration with a 7 bits representation was performed,  best reachable \ac{tdr} would have been 0.171 i.e a relative loss of  41.9 \% of efficiency compared to the optimum \ac{tdr} of 0.298. On the other hand, if network topology was ignored and design space exploration focused only on fixed point data-representation, we show that there can be a loss of 6.8 \% of classification accuracy between two implementations with same data-representation and different topologies (considering our design space size). This underlines how important a holistic approach is, where both topology and data-representation of a \ac{cnn} implementation are explored. 

\begin{figure}[!h]
    \centering
    \includegraphics[width=0.45\textwidth]{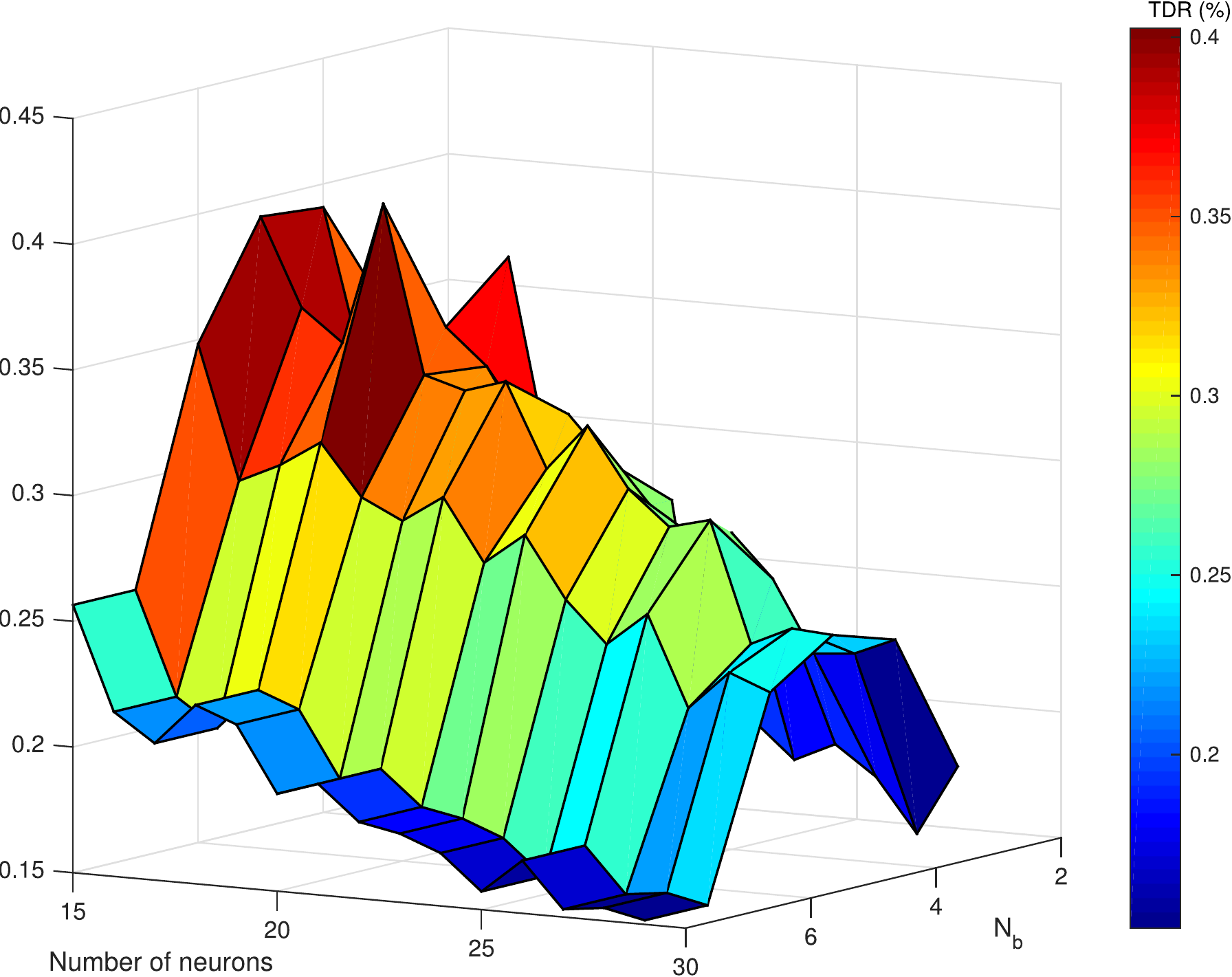}
    \caption{Implementation efficiency of the \ac{cnn} for multiple CNN sizes and fixed point representations }
	\label{holo}
\end{figure}

Figure \ref{holo} gives the result of such exploration as it shows that the most efficient implementations can be obtained after exploring various topologies and data representations. It also appears that a gradient descent optimization can be considered which could lead to a faster exploration process. 

\begin{table}[!h]
\centering
    \begin{tabular}{|c|c|c|c|c|c|c|}
    \hline
    \sc{n$_1$}	&  \sc{n$_2$}	&  \sc{n$_3$}	&   \sc{$B$}	& \sc{tpr}$_{usps}$	&   \sc{dsp} &   \sc{tdr}	\\ \hline
%	3	&	5	&	7	&	5	&	59.4	&	143	&	0.415	\\	\hline
%	4	&	6	&	8	&	5	&	64.8	&	161	&	0.402	\\	\hline
%	3	&	5	&	7	&	4	&	54.7	&	140	&	0.390	\\	\hline
%	4	&	6	&	8	&	3	&	59.1	&	160	&	0.369	\\	\hline
%	3	&	5	&	8	&	5	&	58.4	&	163	&	0.358	\\	\hline
%	3	&	5	&	7	&	6	&	60.6	&	174	&	0.348	\\	\hline
%	4	&	6	&	8	&	4	&	55.6	&	160	&	0.347	\\	\hline
%	3	&	6	&	8	&	5	&	57.4	&	166	&	0.345	\\	\hline
%	3	&	5	&	7	&	3	&	48.2	&	140	&	0.344	\\	\hline
%	5	&	7	&	9	&	5	&	61.2	&	182	&	0.336	\\	\hline
    4	&	8	&	12	&	5	&	73		&	245	&	0.298	\\ \hline 
    5	&	9	&	12	&	5	&	70.1	&	243	&	0.288	\\ \hline 
	4	&	8	&	12	&	6	&	72.5	&	279	&	0.260	\\ \hline
	5	&	9	&	12	&	6	&	70.74	&	275	&	0.257	\\ \hline
	3	&	8	&	13	&	6	&	71.8	&	296	&	0.242	\\ \hline
	4	&	7	&	13	&	6	&	71		&	296	&	0.240	\\ \hline
	3	&	9	&	14	&	6	&	70.4	&	320	&	0.220	\\ \hline
	4	&	8	&	12	&	7	&	73.2	&	428	&	0.171	\\ \hline 
	5	&	9	&	12	&	7	&	71.3	&	417	&	0.171	\\ \hline
	3	&	8	&	13	&	7	&	72.3	&	441	&	0.164	\\ \hline
	4	&	7	&	13	&	7	&	70.8	&	438	&	0.162	\\ \hline
	3	&	9	&	14	&	7	&	71.3	&	475	&	0.150	\\ \hline
    \end{tabular}
    
\caption{Top 10 efficient implementations}
\label{10best}
\end{table}
%or 3D ??

%\newpage

%\newpage
\section{Conclusion}
This work presented a method to optimize \ac{dsp} utilization in \acp{fpga} for \ac{cnn} implementations. It relies on a holistic design space exploration on \ac{cnn} topology and data-representation size to determine the most efficient architecture. As an example, this optimization was applied for \ac{ocr} applications but can be transposed to other \ac{cnn} classification tasks. It has been shown in this paper that a holistic approach is needed to optimize \acp{dsp}, as both fixed point arithmetic and topology network aspects should be explored.  

The soft degradation in terms of quality when the number of bits is reduced or the topology simplified shows that \acp{cnn} are particularly well suited to approximate computing with a controlled rate of errors. Future work aim to improve optimization by augmenting the size of the explored design space and explore the \ac{cnn} depth. Moreover, we expect that more efficient architectures can be implemented when bypassing the Caph \ac{hls} layer and generate directly the corresponding \ac{hdl} of a neural network. Finally, it is planed to transpose design space exploration method to take more hardware constrains into account, such memory or logic elements utilization.     

\section{Acknowledgment}
This work was funded by the french ministry of higher education MESR at Institut Pascal (UMR 6602). We thank them and all the collaborators for their support to this research.

\bibliographystyle{unsrt}
\bibliography{paperbib}
\end{document}